\pdfoutput=1

\documentclass[11pt]{article}

\usepackage[]{ACL2023}

\usepackage{times}
\usepackage{latexsym}

\usepackage[T1]{fontenc}

\usepackage[utf8]{inputenc}

\usepackage{microtype}

\usepackage{inconsolata}

\usepackage{graphicx}
\usepackage{eurosym}
\usepackage{tablefootnote}
\usepackage{kantlipsum}
\usepackage{tabularx}
\graphicspath{ {figures/} }
\usepackage[T1]{fontenc}
\usepackage{kantlipsum}
\usepackage{listings}
\usepackage{stfloats}
\usepackage{csquotes}
\usepackage{array}
\usepackage{booktabs}
\usepackage[super]{nth}
\usepackage[inline,shortlabels]{enumitem}
\usepackage{enumitem}
\usepackage{makecell}
\usepackage{siunitx}
\usepackage{bm}
\usepackage{url}
\usepackage{wrapfig}
\usepackage{amsmath}
\usepackage{amssymb}
\usepackage{mathtools}
\usepackage{amsfonts}
\usepackage{subcaption}
\usepackage{nicefrac}       
\usepackage{tablefootnote}
\usepackage{bbm}
\usepackage{microtype} 
\usepackage{dsfont}
\usepackage{xcolor}
\newcommand{\Dpool}{\mathcal{D}_{\text{pool}}}
\newcommand{\Dlab}{\mathcal{D}_{\text{lab}}}
\newcommand{\Dval}{\mathcal{D}_{\text{val}}}
\newcommand{\Dtest}{\mathcal{D}_{\text{test}}}

\title{On the Limitations of \textit{Simulating} Active Learning}

\author{Katerina Margatina \quad Nikolaos Aletras\\
University of Sheffield\\
\texttt{\{k.margatina, n.aletras\}@sheffield.ac.uk}}

\begin{document}
\maketitle
\begin{abstract}
Active learning (AL) is a \textit{human-and-model-in-the-loop} paradigm that iteratively selects informative unlabeled data for human annotation, aiming to improve over random sampling. However, performing AL experiments with human annotations on-the-fly is a laborious and expensive process, thus unrealistic for academic research. An easy fix to this impediment is to \textit{simulate} AL, by treating an \textit{already} labeled and publicly available dataset as the pool of \textit{unlabeled} data. In this position paper, we first survey recent literature and highlight the challenges across all different steps within the AL loop. We further unveil neglected caveats in the experimental setup that can significantly affect the quality of AL research. We continue with an exploration of how the \textit{simulation} setting can govern  empirical findings, arguing that it might be one of the answers behind the ever posed question \textit{``why do active learning algorithms sometimes fail to outperform random sampling?}''. We argue that evaluating AL algorithms on available labeled datasets might provide a \textit{lower bound} as to their effectiveness in real data. We believe it is essential to collectively shape the best practices for AL research, particularly as engineering advancements in LLMs push the research focus towards data-driven approaches (e.g., data efficiency, alignment, fairness). In light of this, we have developed guidelines for future work. Our aim is to draw attention to these limitations within the community, in the hope of finding ways to address them. 
\end{abstract}

\section{Introduction}\label{sec:intro}
\begin{figure}[t]
    \centering
    \resizebox{0.8\columnwidth}{!}{%
    \includegraphics{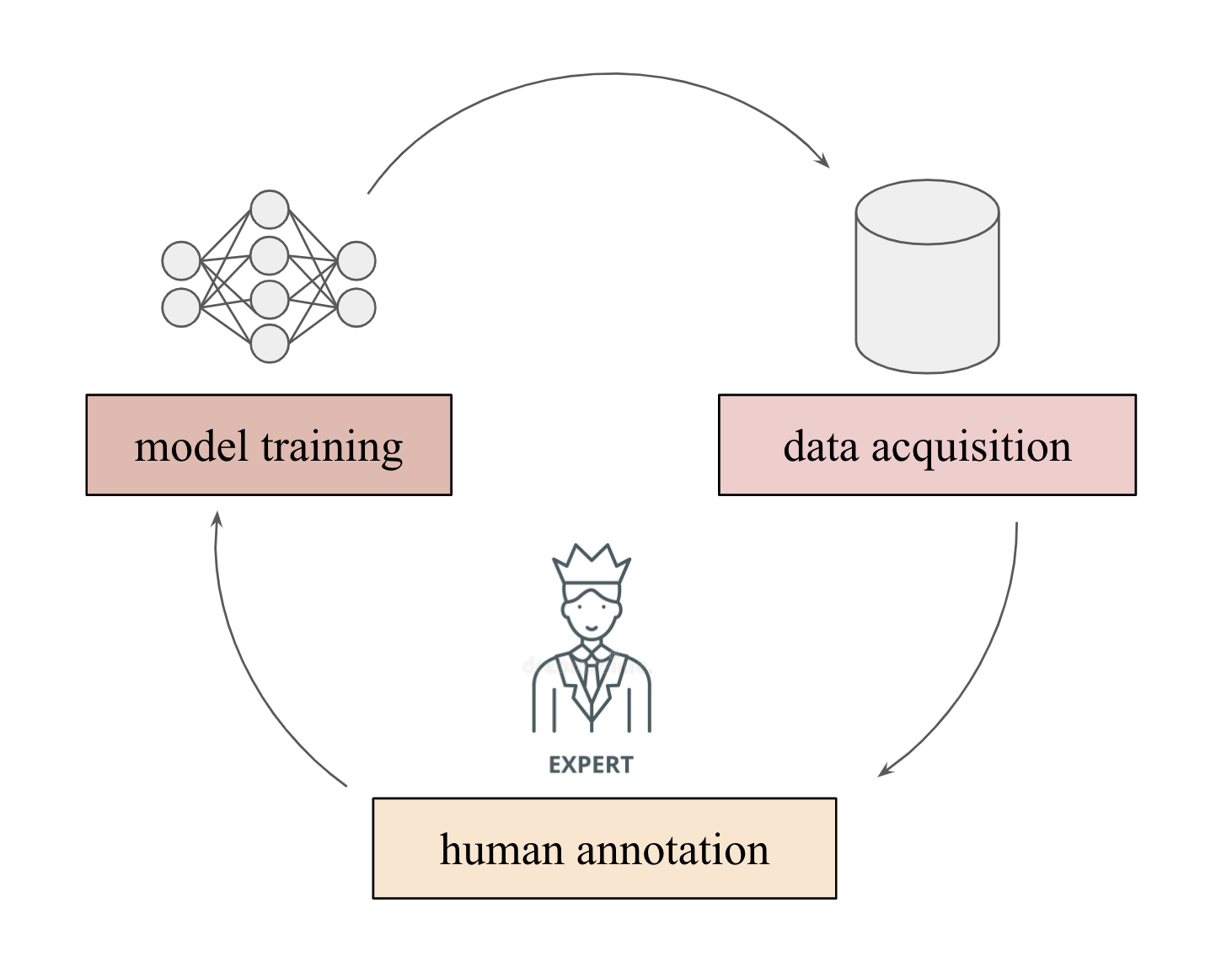}
    }
    \caption{High-level overview of the \textit{train-acquire-annotate} steps of the active learning loop. }
    \label{fig:al_high_level}
\end{figure}

Based on the assumption that ``\textit{not all data is equal}'', active learning (AL)~\cite{Cohn:1996:ALS:1622737.1622744, settles.tr09} aims to identify the most informative data for annotation from a pool (or a stream) of unlabeled data (i.e., data acquisition). With multiple rounds of model training, data acquisition and human annotation (Figure~\ref{fig:al_high_level}), the goal is to achieve \textit{data efficiency}. A data efficient AL algorithm entails that a model achieves satisfactory performance on a held-out test set, by being trained with only a fraction of the acquired data. 

AL has traditionally attracted wide attention in the Natural Language Processing (NLP) community. It has been explored for machine translation~\cite{Haffari2009, Dara2014-vm, Miura2016-gm, zhao-etal-2020-active}, text classification~\cite{ein-dor-etal-2020-active, surveyschroder, margatina-etal-2022-importance,schroder-etal-2023-small}, part-of-speech tagging~\cite{chaudhary-etal-2021-reducing}, coreference~\cite{yuan-etal-2022-adapting} and entity resolution~\cite{Qian2017-xf,kasai-etal-2019-low}, named entity recognition~\cite{erdmann-etal-2019-practical, Shen2017-km, 10.1093/jamia/ocz102}, and natural language inference~\cite{snijders-etal-2023-investigating}, \textit{inter alia}. Still, its potential value is on the rise \cite{aclsurvey_emnlp22}, as the current language model pretraining paradigm continues to advance the state-of-the-art \cite{tamkin2022active}. Under the initial ``not all data is equal'' assumption, it is logical to assume that researchers would try to find the ``most useful'' data to pretrain or adapt their LLMs.

The usual pool-based AL setting is to acquire data from an unlabeled pool, label it, and use it to train a supervised model that, hopefully, obtains satisfactory performance on a test set for the task at hand. This is very similar to the general model-in-the-loop paradigm \cite{karmakharm-etal-2019-journalist, bartolo-etal-2020-beat, bartolo-etal-2022-models, kiela-etal-2021-dynabench, wallace-etal-2022-analyzing}, with the main difference being the AL-based data acquisition stage. The assumption is that, by iteratively selecting data for annotation according to an informativeness criterion, it will result into better model predictive performance compared to randomly sampling and annotate data of the same size.

However, this does not always seem to be the case. A body of work has shown that AL algorithms, that make use of uncertainty~\cite{Lewis:1994:SAT:188490.188495, Cohn:1996:ALS:1622737.1622744,Houlsby2011-qz,pmlr-v70-gal17a}, diversity sampling~\cite{Brinker03incorporatingdiversity, pmlr-v16-bodo11a,conf/iclr/SenerS18} or even more complex acquisition strategies~\cite{DBLP:journals/corr/abs-1802-09841, Ash2020Deep,yuan-etal-2020-cold, margatina-etal-2021-active}, often fail to improve over a simple random sampling baseline~\cite{baldridge-palmer-2009-well,DBLP:journals/corr/abs-1802-09841,lowell-etal-2019-practical,kees-etal-2021-active, karamcheti-etal-2021-mind, snijders-etal-2023-investigating}. Such findings pose a serious question on the practical usefulness of AL, as they do not corroborate its initial core hypothesis that \textit{not all data is equally useful for training a model}. In other words, if we cannot show that one subset of the data is ``better''\footnote{We consider a labeled dataset $A\subset C$ to be ``better'' than a labeled dataset $B\subset C$, both sampled from a corpus C and $|A|=|B|$, if a model $M_A$ trained on $A$ yields higher performance on a test set compared to $M_B$, where both models are identical in terms of architecture, training procedure, etc.}  than another, why do AL in the first place?

Only a small body of work has attempted to explore the pain points of AL. For instance, \citet{karamcheti-etal-2021-mind}, leveraging visualisations from \textit{data maps}~\cite{swayamdipta-etal-2020-dataset}, show that AL algorithms tend to acquire \textit{collective outliers} (i.e. groups of examples that deviate from the rest of the examples but cluster together), thus explaining the utter failure of eight AL algorithms to outperform random sampling in visual question answering. Building on this work, more recently \citet{snijders-etal-2023-investigating} corroborate these findings for the task of natural language inference and further show that uncertainty based AL methods recover and even surpass random selection when hard-to-learn data points are removed from the pool.
\citet{lowell-etal-2019-practical} show that the benefits of AL with certain models and domains do not generalize reliably across models and tasks. This could be problematic since, in practice, one might not have the means to explore and compare alternative AL strategies. They also show that an actively acquired dataset using a certain model-in-the-loop, may be disadvantageous for training models of a different family, raising the issue of whether the downsides inherent to AL are worth the modest and inconsistent performance gains it tends to afford.

In this paper, we aim to explore all possible limitations that researchers and practitioners currently face when doing research on AL \cite{aclsurvey_emnlp22}. We first describe the process of pool-based AL (Figure~\ref{fig:al_high_level}) and identify challenges in every step of the iterative process (\S \ref{sec:challenges}). Next, we unearth obscure details that are often left unstated and under-explored (\S \ref{sec:fineprint}). We then delve into a more philosophical discussion of the role of simulation and its connection to real practical applications (\S \ref{sec:simulation}). Finally, we provide guidelines for future work (\S \ref{sec:future}) and conclusions (\S \ref{sec:conclusion}), aspiring to promote neglected, but valuable, ideas to improve the direction of research in active learning.

\section{Challenges in the Active Learning Loop}
\label{sec:challenges}
We first introduce the typical steps in the pool-based AL setting \cite{Lewis:1994:SAT:188490.188495} and identify several challenges that an AL practitioner has to deal with, across all steps (Figure~\ref{fig:al_loop}).\footnote{We point the reader to the comprehensive survey of \citet{aclsurvey_emnlp22} for a more in-depth exploration of recent literature in AL.} 

\subsection{Problem Definition}\label{sec:setting}
Consider the experimental scenario where we want to model a specific NLP task for which we do not yet have any labeled data, but we have access to a large pool of unlabeled data $\Dpool$. We assume that it is unrealistic (e.g., laborious, expensive) to have humans annotating all of it. $\Dpool$ constitutes the textual corpus from which we want to sample a fraction of the most \textit{useful} (e.g., informative, representative) data points for human annotation. In order to perform active learning, we need an initial labeled dataset $\Dlab$, often called ``seed'' dataset, to be used for training a task-specific model with supervised learning. To evaluate the model, we need a usually small validation set for model selection $\Dval$ and a held out test set $\Dtest$ to evaluate the model's generalization. We use $\Dlab$ and $\Dval$ to train the first model and then test it on $\Dtest$. 

In this stage, we start acquiring labeled data for model training. Data points are sampled from $\Dpool$ via an acquisition strategy and subsequently passed to human annotators for labeling. The acquisition function selects a batch of data $Q\subset\Dpool$ according to some informativeness criterion and can either use the model-in-the-loop or not.  We employ crowdsourcing or expert annotators to label the selected batch $Q$ which then is appended to the labeled dataset $\Dlab$. 

Now that we have augmented the seed dataset with more data, we re-train the model on the new training dataset, $\Dlab$. We test the new model on $\Dtest$ and we stop if we obtain satisfactory performance or if the budget for annotation has run out (or using any other stopping criterion). If we do not want to stop, we use the acquisition function to select more unlabeled data from $\Dpool$, which we annotate and append to $\Dlab$, etc. This is the AL loop shown in Figure~\ref{fig:al_loop}.

\subsection{Active Learning Design}\label{sec:init_steup}
\paragraph{Seed dataset}
We start the AL loop (\S\ref{sec:setting}) by defining an initial labeled ``seed dataset'' (Figure~\ref{fig:al_loop}: \fbox{1}). The seed dataset plays an important role, as it will be used to train the the first model-in-the-loop \cite{tomanek-etal-2009-proper,horbach-palmer-2016-investigating}. In AL research, we typically address the cold-start problem by sampling from $\Dpool$ with a uniform distribution for each class, either retaining the true label distribution or choosing data that form a balanced label distribution.\footnote{In AL research, a fully labeled dataset is typically \textit{treated} as an \textit{unlabeled} $\Dpool$ by entirely ignoring its labels, while in reality we \textit{do} have access to them. Hence, the labels implicitly play a role in the design of the AL experiment. We analyze our criticism to this seemingly ``random sampling'' approach to form the seed dataset in \S\ref{sec:upper_bound}.} This is merely a convenient design choice, as it is simple and easy to implement. However, sampling the seed dataset this way, does not really reflect a real-world setting where the label distribution of the (unlabeled data of the) pool is actually unknown. 

\citet{Prabhu2019-hm} performed a study of such sampling bias in AL, showing no effect in different seed datasets across the considered methods. \citet{ein-dor-etal-2020-active} also experimented with different imbalanced seed datasets, showing that AL improves over random sampling in settings with highest imbalance.
 
Furthermore, the choice of the seed dataset has a direct effect on the entire AL design because the first model-in-the-loop marks the reference point of the performance in $\Dtest$. In other words, the performance of the first model is essentially the baseline, according to which a practitioner will plan the AL loop based on the goal performance and the available budget. It is thus essential to revisit existing approaches on choosing the seed dataset \cite{Kang2004UsingCS,vlachos-2006-active,FLAIRS101305,yuan-etal-2020-cold} and evaluate them towards a realistic simulation of an AL experiment.

\paragraph{Number of iterations \& acquisition budget}
After choosing the seed dataset it is natural to decide the number of iterations, the acquisition size (the size of the acquired batch $\mathcal{Q}$) and the budget (the size of the actively collected $\Dlab$) of the AL experiment. This is another part where literature does not offer concrete explanations on the design choice. Papers that address the cold-start problem would naturally focus on the very few first AL iterations~\cite{yuan-etal-2020-cold}, while others might simulate AL until a certain percentage of the pool has been annotated~\cite{Prabhu2019-hm,lowell-etal-2019-practical,zhao-etal-2020-active,zhang-plank-2021-cartography-active,margatina-etal-2022-importance} or until a certain fixed and predefined number of examples has been annotated~\cite{ein-dor-etal-2020-active,kirsch-test-dist}.

\subsection{Model Training}
\begin{figure*}[t]
    \centering
    \resizebox{0.9\textwidth}{!}{%
    \includegraphics{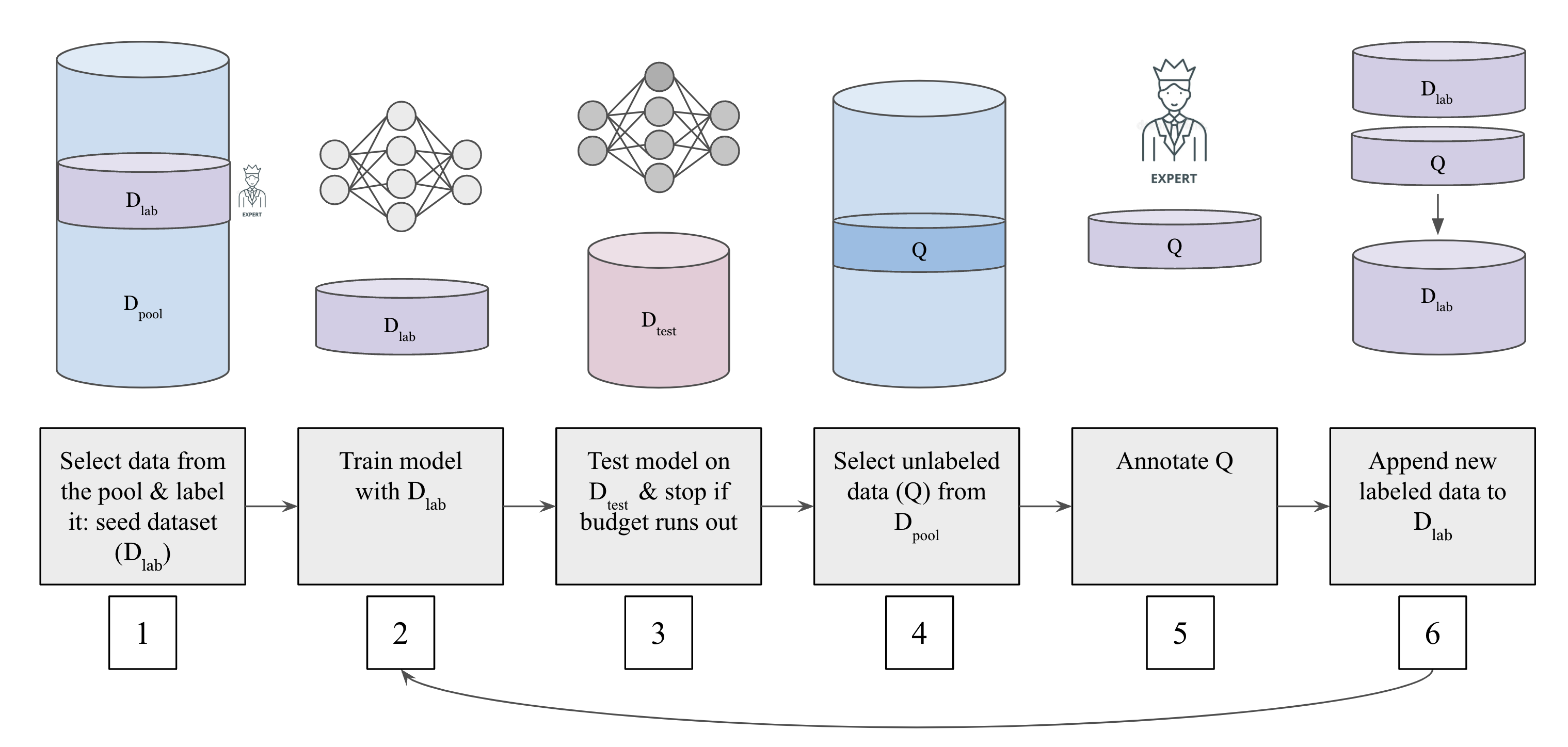}
    }
    \caption{Distinct steps of the active learning loop ($1$--$6$). We use  \textcolor{blue}{blue} for the unlabeled data,  \textcolor{purple}{purple} for the labeled data and  \textcolor{red}{red} for the (labeled) test data. }
    \label{fig:al_loop}
\end{figure*}

We now train the model-in-the-loop with the available labeled dataset $\Dlab$ (Figure~\ref{fig:al_loop}: \fbox{2}). Interestingly, there are not many studies that explore how we should properly train the model in the low data resource setting of AL. Existing approaches include semi-supervised learning \cite{10.5555/645527.757765,tomanek-hahn-2009-semi,dasgupta-ng-2009-mine,yu-etal-2022-actune}, weak supervision \cite{Ni2019MergingWA,qian-etal-2020-learning,brantley-etal-2020-active,zhang-etal-2022-prompt} and data augmentation \cite{zhang-etal-2020-seqmix, zhao-etal-2020-active,hu-neubig-2021-phrase}, with the most prevalent approach currently to be transfer learning from pretrained language models \cite{ein-dor-etal-2020-active,margatina-etal-2021-active,tamkin2022active}. Recently, \citet{margatina-etal-2022-importance} showed large performance gains by adapting the pretrained language model to the task using the unlabeled data of the pool (i.e., task adaptive pretraining by~\citet{gururangan-etal-2020-dont}). The authors also proposed an adaptive fine-tuning technique to account for the varying size of $\Dlab$ showing extra increase in $\Dtest$ performance.

Still, there is room for improvement in this rather under-explored area. Especially now, state-of-the-art NLP pretrained language models consist of many millions or even billions of parameters. In AL we often deal with a small $\Dlab$ of a few hundred examples, thus adapting the training strategy is not trivial.

\subsection{Data Acquisition}\label{sec:acquisition}
The data acquisition step (Figure~\ref{fig:al_loop}: \fbox{4}) is probably the core of the AL process and can be performed in various ways.\footnote{In literature, the terms \textit{data selection method}, \textit{query strategy} and \textit{acquisition function} are often used interchangeably. }

\citet{aclsurvey_emnlp22} provide a thorough literature review of query strategies, dividing them into two broad families. The first is based on \textit{informativeness}, and methods in this family treat each candidate instance individually, assign a score and select the top (or bottom) instances based on the ranking of the scores. Major sub-categories of methods that belong in the informativeness family are uncertainty sampling~\cite{Lewis:1994:SAT:188490.188495,Culotta05reducinglabeling, zhang-plank-2021-cartography-active,schroder-etal-2022-revisiting}, divergence-based algorithms~\cite{DBLP:journals/corr/abs-1802-09841, margatina-etal-2021-active,zhang-etal-2022-allsh}, disagreement-based~\cite{10.1145/130385.130417,Houlsby2011-qz,pmlr-v70-gal17a,siddhant-lipton-2018-deep,Kirsch2019-lk,zeng-zubiaga-2023-active}, gradient-based \cite{NIPS2007_a1519de5,settles-craven-2008-analysis} and performance prediction \cite{Roy01towardoptimal,Konyushkova2017-jj,Bachman2017-wg,
liu-etal-2018-learning-actively}.

The second family is representativeness and takes into account how instances of the pool correlate with each other, in order to avoid sampling bias harms from treating each instance individually. Density-based methods choose the most representative instances of the unlabeled pool \cite{Ambati2010-hk,zhao-etal-2020-active,zhu-etal-2008-active}, while others opt for discriminative data points that differ from the already labeled dataset \cite{Gissin2019-os, erdmann-etal-2019-practical}. A commonly adopted category in this family is batch diversity, where algorithms select a batch of diverse data points from the pool at each iteration~\cite{Brinker03incorporatingdiversity, pmlr-v16-bodo11a,zhu-etal-2008-active,DBLP:journals/corr/abs-1711-00941,Zhdanov2019-mg,yu-etal-2022-actune}, with core-set \cite{conf/iclr/SenerS18} to be the most common approach.

Naturally, there are hybrid acquisition functions that combine informativeness and representativeness \cite{yuan-etal-2020-cold, Ash2020Deep,shi-etal-2021-diversity}.
Still, among the aforementioned methods there is not a universally superior acquisition function that consistently outperforms all others. Thus, which data to acquire is an active area of research.

\subsection{Data Annotation}
After selecting a subset $Q$ from $\Dpool$ with an acquisition function, we send the acquired unlabeled data to humans for annotation (Figure~\ref{fig:al_loop}: \fbox{5}). In the simulation AL setting, we do not focus at this part at all, as we \textit{already} have the labels of the actively acquired batch. However, a question that naturally arises is: \emph{Are all examples equally easy to annotate?} In simulation, all instances take equally long to label. This does not account for the fact that hard instances for the classifier are often hard for humans as well \cite{hachey-etal-2005-investigating,baldridge-osborne-2004-active}, therefore the current experimental setting is limiting and research for cost-aware selection strategies \cite{10.1145/1458082.1458165, tomanek-hahn-2010-comparison,10.1093/jamia/ocz102} is required. This would include explicit exploration of the synergies between random or actively acquired data and annotator expertise~\cite{baldridge-palmer-2009-well}.

\subsection{Stopping Criterion}
Finally, another active area of research is to develop effective methods for stopping AL (Figure~\ref{fig:al_loop}: \fbox{3}). In simulation, we typically decide as a budget a number of examples or a percentage of $\Dpool$ up to which we ``aford'' to annotate. However, in both research and real world applications, it is not clear if the model performance has reached a plateau. The stopping criterion should not be pre-defined by a heuristic, but rather a product of a well-designed experimental setting \cite{VLACHOS2008295,tomanek-hahn-2010-comparison,pmlr-v108-ishibashi20a,pullsrstrecker,DBLP:journals/corr/abs-2202-02794,DBLP:journals/corr/abs-2201-05460}.\footnote{Unless of course the actual budget is spent, where in real world settings this is effectively the stopping criterion.}

\section{The Fine Print}\label{sec:fineprint}
Previously, we presented specific challenges across different steps in the AL loop that researchers and practitioners need to address. Still, these challenges have long been attracting the attention of the research community. 
Interestingly, there are more caveats, that someone with no AL experience might have never encountered or even imagined. Hence, in this section we aim to unveil several such small details that still remain unexplored. 

\subsection{Hyperparameter Tuning}\label{sec:hyp_tuning}
A possibly major issue of the current academic status quo in AL, is that researchers often do not tune the models-in-the-loop. This is mostly due to limitations related to time and compute constrains. For instance, a paper that proposes a new acquisition function would be required to run experiments for multiple baselines, iterations, random seeds and datasets. For example, a modest experiment including $a=5$ acquisition functions, $i=10$ AL iterations, $n=5$ random seeds and $d=5$ datasets, would reach an outstanding number of minimum $a\times i\times n\times d=1,250$ trained models in total. This makes it rather hard to perform hyperparameter tuning of all these models in every AL loop, so it is the norm to use the same model architecture and hyperparameters to train all models.

In reality, practitioners that want to use AL, apply it \textit{once}. Therefore, they most likely afford to tune the one and only model-in-the-loop. The question that arises then, is ``\textit{do the findings of AL experiments that do not tune the models generalize to scenarios where all models-in-the-loop are tuned}''? In other words, if an AL algorithm $A$ performs better than $B$ according to an experimental finding, would this be the case if we applied hyperparameter tuning to the models of both algorithms? Wouldn't it be possible that, with another configuration of hyperparameters, $B$ performed better in the end?

\subsection{Model Stability}\label{sec:crash}
In parallel, another undisclosed detail is what researchers do when the models-in-the-loop are unstable (i.e., \textit{crash}). This essentially means that for some reason the optimisation of the model might fail and the model never converges leading to extremely poor predictive performance. Perhaps before the deep learning era such a problem did not exist, but now it is a likely phenomenon.

\citet{https://doi.org/10.48550/arxiv.2002.06305}  showed that many fine-tuning experiments diverged part of the way through training especially on small datasets. AL is by definition connected with low-data resource settings, as the gains of data efficiency are meaningful in the scenario when labeled data is scarce. 

In light of this challenge, there is no consensus as to what an AL researcher or practitioner should do to alleviate this problem. One can choose to re-train the model with a different random seed, or do nothing. Though, it is non-trivial under which condition one should choose to re-train the model, since it is common that not always test performance improves from one AL iteration to the next. 

Furthermore, there is currently no study that explores how much AL algorithms, that use the model-in-the-loop for acquisition, suffer by this problem. For instance, consider an uncertainty-based AL algorithm that uses the predictive probability distribution of the model to select the most uncertain data points from the pool. If the model crashes, then its uncertainty estimates are not meaningful, thus the data acquisition function does not work as expected. In effect, the sampling method turns to a uniform distribution (i.e., the random sampling baseline).

\subsection{Active Learning Evaluation}\label{sec:al_eval}
Another important challenge is the evaluation framework for AL. Evaluating the \textit{actual} contribution of an AL method against its competitors would require to perform the same iterative \textit{train-acquire-annotate} experiment (Figure~\ref{fig:al_high_level}) for all AL methods in the exact same data setting and with real human annotations. Certainly, such a laborious and expensive process is prohibitive for academic research, which is why we perform simulations by treating an \textit{already} labeled and open-source dataset as a pool of unlabeled data. 

Still, even if we were able to perform the experiments in real life, it is not trivial how to properly define when one method is better than another. This is because AL experiments include multiple rounds of annotation, thus multiple trained models and multiple scores in the test set(s). In cases with no clear difference between the algorithms compared, how should we do a fair comparison?

Previous work presents tables comparing the test set performance of the last model, often ignoring performance in previous loops~\cite{Prabhu2019-hm,mussmann2020importance}. The vast majority of previous work though uses plots to visualize the performance over the AL iterations~\cite{lowell-etal-2019-practical,ein-dor-etal-2020-active} and in some cases offer a more detailed visualization with the variance due to the random seeds~\cite{yuan-etal-2020-cold,kirsch-test-dist, margatina-etal-2021-active}.

\subsection{The Test of Time}\label{sec:test_of_time}
\citet{settles.tr09} eloquently defines the ``test of time'' problem that AL faces: ``\textit{A training set built in cooperation with an active learner is inherently tied to the model that was used to generate it (i.e., the class of the model selecting the queries). Therefore, the labeled instances are a biased distribution, not drawn i.i.d. from the underlying natural density. If one were to change model classes—as we often do in machine learning when the state of the art advances—this training set may no longer be as
useful to the new model class}''.

Several years later, in the deep learning era, \citet{lowell-etal-2019-practical} indeed corroborates this concern. They demonstrate that a model from a certain family (e.g., convolution neural networks) might perform better when trained with a random subset of a pool, than an actively acquired dataset with a model-in-the-loop of a different family (e.g., recurrent neural networks). Related to the ``test of time'' challenge, it is rarely investigated whether the training data actively acquired with one model will confer benefits if used to train a second model (as compared to randomly sampled data from the same pool). Given that datasets often outlive learning algorithms, this is an important practical consideration \cite{baldridge-osborne-2004-active,lowell-etal-2019-practical, shelmanov-etal-2021-active}.

\section{Active Learning in \textit{Simulated} vs. Real World Settings}\label{sec:simulation}
\begin{quote}
\textit{Is it truly logical to consider an already cleaned (preprocessed), typically published open-source labeled dataset as an unlabeled data pool for pool-based active learning simulation, with the expectation that any conclusions drawn will be applicable to real-world scenarios?}

\end{quote}

The convenience and scalability of simulation make it an undoubtedly appealing approach for advancing machine learning research. In NLP, 
when tackling a specific task, for instance summarization, researchers often experiment with the limited availability of labeled summarization datasets, aiming to gain valuable insights and improve summarization models across various domains and languages.
While this approach may not be ideal, it is a practical solution.  \textit{What makes the sub-field of active learning different?} 

Admittedly, progress has, and will be made in AL research by leveraging simulation environments, similar to other areas within machine learning. Thus, there is no inherent requirement for a radically different approach in AL. We believe that simulating AL is indispensable for developing new methods and advancing the state-of-the-art.

Nonetheless, we argue that a slight distinction should be taken into account. AL is an iterative process that aims to obtain the smallest possible amount of labeled \textit{data} given a substantially larger pool of unlabeled data for maximizing predictive performance on a given task. The significant difference between developing models and constructing datasets lies in the fact that if a model is poorly trained, it can simply be retrained.
Conversely, in AL, there exists a finite budget for acquiring annotations, and once it is expended, \textit{there is no going back}. Consequently, we must have confidence that the AL state-of-the-art established through research simulations will perform equally well in practical applications.


Given these considerations, we advocate for a more critical approach to conducting simulation AL experiments. 
We should be addressing all the challenges (\S \ref{sec:challenges}) and the experimental limitations (\S \ref{sec:fineprint}) discussed previously, while acknowledging the disparities between the simulation environment and real-world applications (\S \ref{sec:lower_bound}). Given that datasets tend to outlast models~\cite{lowell-etal-2019-practical}, we firmly believe that it is crucial to ensure the trustworthiness of AL research findings and their generalizability to real-world active data collection. This will contribute to the generation of high-quality datasets that stand the test of time (\S \ref{sec:test_of_time}). 

\subsection{Simulation as a \textit{Lower} Bound of Active Learning}\label{sec:lower_bound}
The distribution gap between benchmark datasets in common ML tasks and data encountered in a real world production setting is well known~\cite{Bengio2020A, pmlr-v139-koh21a, Wang_2018,https://doi.org/10.48550/arxiv.2111.10497}. 

\paragraph{High Quality Data} It is common practice for researchers to carefully curate the data to be labeled properly, often collecting multiple human annotations per example and discarding instances with disagreeing labels. When datasets are introduced in papers published in prestigious conferences or journals, it is expected that they should be of the highest quality, with an in-depth analysis of its data collection procedure, label distribution and other statistics. Nonetheless, it is important to acknowledge that such datasets may not encompass the entire spectrum of language variations encountered in real-world environments~\cite{https://doi.org/10.48550/arxiv.2111.10497}. Consequently, it remains uncertain whether an AL algorithm would generalize effectively to unfiltered raw data. Specifically, we hypothesize that the filtered data would be largely \textit{more homogeneous} than the initial ``pool''. Assuming that the simulation $\Dpool$ is a somewhat homogeneous dataset, we can expect that \textit{any} subset of data points drawn from it would, consequently, be more or less identical.\footnote{Here we do not hint that all textual instances of a dataset are actually identical, but that they are more similar between them compared to the larger pool that they were created from.} Therefore, if we train a model in each such subset, we would expect to obtain similar performance on test data due to the similarity between the training sets. From this perspective, random (uniform) sampling from a homogeneous pool can be considered a rudimentary form of diversity sampling. 

\paragraph{Low Quality Data} 
In contrast, it is possible that a publicly available dataset used for AL research  may contain data of inferior quality, characterized by outliers such as repetitive instances, inadequate text filtering, incorrect labels, and implausible examples, among others. 
In such cases, an AL acquisition strategy, particularly one based on model uncertainty, may consistently select these instances for labeling due to their high level of data difficulty and uncertainty. 
Previous studies~\cite{karamcheti-etal-2021-mind,snijders-etal-2023-investigating} have demonstrated the occurrence of this phenomenon, which poses a significant challenge as it undermines the potential value of AL.
In a real-world AL scenario, it is plausible to have a dedicated team responsible for assessing the quality of acquired data and discarding instances of subpar quality. However, within the confines of a simulation, such data filtering is typically absent from the researcher's perspective, leading to potentially misleading experimental outcomes. \citet{snijders-etal-2023-investigating} tried to address this issue in a multi-source setting for the task of natural language inference, and showed that while uncertainty-based strategies perform poorly due to the acquisition of collective outliers, when outliers are removed (from the pool), AL algorithms exhibited a noteworthy recovery and outperformed random baselines.

\subsection{Simulation as an \textit{Upper} Bound of Active Learning}\label{sec:upper_bound}
However, one might argue for the exact opposite.
\paragraph{Favored Design Choices}  Previously, we mentioned that when selecting the seed dataset (\S \ref{sec:init_steup}) we typically randomly sample data from $\Dpool$, while keeping the label distribution of the true training set.\footnote{The ``true training set'' is the original one used as the pool ($\Dpool$) by removing the labels.} Hence, 
a balanced seed dataset is typically obtained, given that most classification datasets tend to exhibit a balanced label distribution. 
In effect, the label distribution of $\Dpool$ would also be balanced, setting a strict constraint on the AL simulation setting, as the actual label distribution of the unlabeled data should in reality be \textit{unknown}. In other words, such subtle choices in the experimental design can introduce bias, making the simulated settings more trivial than more challenging real world AL settings where there is uncertainty as to the quality and the label distribution of data crawled online, that typically constitute the unlabeled pool.

\paragraph{Temporal Drift \& Model Mismatch}
Datasets intended for research purposes are often constructed within a fixed timeframe, with minimal consideration for temporal concept drift issues~\cite{rottger-pierrehumbert-2021-temporal-adaptation, lazaridou2021mind, margatina-etal-2023-dynamic}. However, it is important to recognize that this may not align with real-world applications, where the data distribution undergoes changes over time. The utilization of random and standard splits, commonly employed in AL research, can lead to overly optimistic performance estimates \cite{sogaard-etal-2021-need}, which may not generalize to the challenges presented by real-world scenarios.
Consequently, practitioners should consider this limitation when designing their active learning  experiments. \citet{lowell-etal-2019-practical} also raises several practical obstacles neglected in AL research, such as that the acquired dataset may be disadvantageous for training subsequent models, and concludes that academic investigations of AL typically omit key real-world considerations that might overestimate its utility.

\subsection{Main Takeaway}
In summary, there exist compelling arguments that support both perspectives: simulation can serve as a lower bound by impeding the true advancement of AL methods, or it can implicitly favor AL experimental design, thus providing an upper bound for evaluation. 
The validity of these arguments likely varies across different cases. We can claim with certainty that this simulation setting, as described in this paper, is a far from perfect framework to evaluate AL algorithms among them and against random sampling. Nevertheless, we hypothesize that the lower bound argument (\S \ref{sec:lower_bound}) might be more truthful. 
It is conceivable that AL data selection approaches may exhibit similar performance levels, either due to a lack of variation and diversity in the sampled pool of data or due to the presence of outliers that are not eliminated during the iterations. Hence, we contend that \textit{simulation can be perceived as a lower bound for AL performance}, which helps explain why AL methods struggle to surpass the performance of random sampling.
We undoubtedly believe that we can only obtain such answers by \textit{exploring the AL simulation space in depth and by performing thorough analysis and extensive experiments to contrast the two theories.}

\subsection{Active Learning in the LLMs Era}
The field of active learning holds considerable importance in the context of the current era of Large Language Models (LLMs). AL is inherently intertwined with data-driven approaches that underpin recent advancements in artificial intelligence, such as reinforcement learning from human feedback (RLHF)~\cite{christiano2023deep,chatgpt,openai2023gpt4,bai2022training}. AL and RLHF represent two distinct approaches that tackle diverse aspects of the overarching problem of AI alignment~\cite{askell2021general}. AL primarily focuses on optimizing the data acquisition process by selectively choosing informative instances for labeling, primarily within supervised or semi-supervised learning paradigms. On the other hand, RLHF aims to train reinforcement learning agents by utilizing human feedback as a means to surmount challenges associated with traditional reward signals. Despite their disparate methodologies, both AL and RLHF emphasize the criticality of incorporating human involvement to enhance the performance of machine learning and AI systems. Through active engagement of humans in the training process, AL and RLHF contribute to the development of AI systems that exhibit greater alignment with human values and demonstrate enhanced accountability~\cite{bai2022training,bai2022constitutional,ganguli2022red,glaese2022improving,sun2023principledriven}. Consequently, the synergistic relationship between these two approaches warrants further exploration, as it holds the potential to leverage AL techniques in order to augment the data efficiency and robustness of RLHF methods.
%

\section{Guidelines for Future Work}\label{sec:future}
Given the inherent limitations of simulated AL settings, we propose  guidelines to improve trustworthiness and robustness in AL research. 

\paragraph{Transparency} Our first recommendation is a call for transparency, which essentially means to \textit{report everything} \cite{dodge-etal-2019-show}. Every detail of the experimental setup, the implementation and the results, would be extremely helpful to properly evaluate the soundness of the experiments. We urge AL researchers to make use of the Appendix (or other means such as more detailed technical reports) to communicate interesting (or not) findings and problems, so that all details (\S \ref{sec:fineprint}) are accessible.

\paragraph{Thorough Experimental Settings} We also hope to incentivize researchers to \textit{properly think about their experimental settings}, with a focus on ethical and practical considerations. We argue that it is important to compare as many algorithms as possible, aiming to have results and findings that generalize across datasets, tasks and domains.
Moreover, we endorse research endeavors that aim to simulate more realistic settings for active learning, such as exploration of AL  across multiple domains~\cite{longpre-multi-domain,snijders-etal-2023-investigating}. Additionally, we advocate for investigations into active learning techniques for languages beyond English, as the prevailing body of research predominantly focuses on English datasets~\cite{Bender2011OnAA}.

\paragraph{Evaluation Protocol} 
We strongly encourage researchers to prioritize the establishment of fair comparisons among different methods and to provide thorough and extensive presentation of results, including the consideration of variance across different random seeds, in order to ensure robustness and reliability of findings. Generally, we argue that there is room for improvement of the active learning evaluation framework and we should explore approaches from other fields that promote more rigorous experimental and evaluation frameworks \cite{artetxe-etal-2020-call}.

\paragraph{Analysis} We place additional emphasis on the essential requirement of conducting comprehensive analysis of active learning results. It is imperative to delve into the nuances of how different AL algorithms diverge and the extent of similarity (or dissimilarity) among the actively acquired datasets. It is incumbent upon AL research papers to extend beyond the results section and include an extensive analysis component, which provides deeper insights and understanding, as in \citet{
ein-dor-etal-2020-active, yuan-etal-2020-cold, margatina-etal-2021-active, zhou2021towards, snijders-etal-2023-investigating}, among others. If we aim to unveil why an AL algorithm fails to outperform another (or the random baseline), we need to understand which data it selected in the first place, and why.

\paragraph{Reproducibility} The reproducibility of active learning  experiments can be challenging due to the complex nature of a typical AL experiment, involving multiple rounds of model training and evaluation, which can be computationally demanding. However, we strongly advocate for AL practitioners and researchers to prioritize the release of their codebase and provide comprehensive instructions for future researchers aiming to build upon their work. By making the code and associated resources available, the research community can foster transparency, facilitate replication, and enable further advancements in AL methodologies. 

\paragraph{Efficiency}
In addition, we propose the release of actively acquired datasets generated by different AL algorithms, which would greatly contribute to research focused on the data-centric and interpretability aspects of AL. Particularly in the context of utilizing AL with large-scale models, it becomes crucial to establish the actively acquired data from other studies as baselines, rather than re-running the entire process from the beginning. Such an approach would not only enhance transparency, but also promote efficiency and eco-friendly practices within the research community.

\section{Conclusion}\label{sec:conclusion}
In this position paper, we examine the numerous challenges encountered throughout the various stages of the active learning pipeline. Additionally, we provide a comprehensive overview of the often-overlooked limitations within the AL research community, with the intention of illuminating obscure experimental design choices. Furthermore, we delve into a thorough exploration of the limitations associated with simulation in AL, engaging in a critical discussion regarding its potential as either a lower or upper bound on AL performance. Lastly, we put forth guidelines for future research directions, aimed at enhancing the robustness and credibility of AL research for effective real-world applications. 
This perspective is particularly timely, particularly considering the notable advancements in modeling within the field NLP (e.g., ChatGPT\footnote{\url{https://openai.com/blog/chatgpt}}, Claude\footnote{\url{https://www.anthropic.com/index/introducing-claude}}) . These advancements have resulted in a shift of emphasis towards a more data-centric approach in machine learning research, emphasizing the significance of carefully selecting relevant data to enhance models and ensure their alignment with human values.

\section*{Limitations}
In this position paper, we have strived to provide a comprehensive overview, acknowledging that there may be relevant research papers that have inadvertently escaped our attention. While we have made efforts to include a diverse range of related work from various fields, such as machine learning and computer vision, it is important to note that our analysis predominantly focuses on AL papers presented at NLP conferences. Moreover, it is worth mentioning that the majority, if not all, of the AL papers examined and referenced in this survey are centered around the English language, thereby limiting the generalizability and applicability of our findings and critiques to other languages and contexts. 
We wish to emphasize that the speculations put forth in this position paper carry no substantial risks, as they are substantiated by peer-reviewed papers, and our hypotheses (\S \ref{sec:simulation}) are explicitly stated as such, representing conjectures rather than definitive findings regarding the role of simulation in AL research.
We sincerely hope that this paper stimulates robust discussions and undergoes thorough scrutiny by experts in the field, with the ultimate objective of serving as a valuable guideline for AL researchers, particularly graduate students, seeking to engage in active learning research. Above all, \textit{we earnestly urge researchers equipped with the necessary resources to conduct experiments and analyses that evaluate our hypotheses, striving to bridge the gap between research and real-world settings in the context of active learning.}

\section*{Acknowledgements}
We would like to thank the anonymous reviewers for their insightful feedback. Both authors are supported by an Amazon Alexa Fellowship.

\bibliography{custom}
\bibliographystyle{acl_natbib}




\end{document}